\documentclass[10pt,twocolumn,letterpaper]{article}

\usepackage{iccv}
\usepackage{times}
\usepackage{epsfig}
\usepackage{graphicx}
\graphicspath{{Figures/}}
\usepackage{amsmath}
\usepackage{amssymb}
\usepackage{bbm}

\usepackage{algorithm} 
\usepackage{algpseudocode}

\usepackage{stfloats}

\usepackage[breaklinks=true,bookmarks=false]{hyperref}

\iccvfinalcopy 

\def\iccvPaperID{****} 

\ificcvfinal\fi

\begin{document}

\title{Contrastive Learning and Self-Training for Unsupervised Domain Adaptation in Semantic Segmentation}

\author{Robert A. Marsden, Alexander Bartler, Mario D\"obler, Bin Yang\\
Institute of Signal Processing and System Theory, University of Stuttgart, Germany\\
{\tt\small \{robert.marsden, alexander.bartler, mario.doebler, bin.yang\}@iss.uni-stuttgart.de}
}

\maketitle
\ificcvfinal\thispagestyle{empty}\fi

\begin{abstract}
Deep convolutional neural networks have considerably improved state-of-the-art results for semantic segmentation. Nevertheless, even modern architectures lack the ability to generalize well to a test dataset that originates from a different domain. To avoid the costly annotation of training data for unseen domains, unsupervised domain adaptation (UDA) attempts to provide efficient knowledge transfer from a labeled source domain to an unlabeled target domain. Previous work has mainly focused on minimizing the discrepancy between the two domains by using adversarial training or self-training. While adversarial training may fail to align the correct semantic categories as it minimizes the discrepancy between the global distributions, self-training raises the question of how to provide reliable pseudo-labels. To align the correct semantic categories across domains, we propose a contrastive learning approach that adapts category-wise centroids across domains. Furthermore, we extend our method with self-training, where we use a memory-efficient temporal ensemble to generate consistent and reliable pseudo-labels. Although both contrastive learning and self-training (CLST) through temporal ensembling enable knowledge transfer between two domains, it is their combination that leads to a symbiotic structure. We validate our approach on two domain adaptation benchmarks: GTA5 $\rightarrow$ Cityscapes and SYNTHIA $\rightarrow$ Cityscapes. Our method achieves better or comparable results than the state-of-the-art. We will make the code publicly available.
   
\end{abstract}

\section{Introduction}
The goal in semantic image segmentation is to assign the correct class label to each pixel. This makes it suitable for complex image-based scene analysis that is required in applications like automated driving. However, in order to generate a labeled training dataset, pixel-level annotation must first be performed by humans. Since detailed manual labeling can take 90 minutes per image \cite{Cityscapes}, it is associated with high costs. A potential workaround would be to generate the images and corresponding segmentation maps synthetically using computer game environments like Grand Theft Auto V (GTA5) \cite{GTA5}. However, even current segmentation models \cite{Deeplabv3+, HRNet, Deeplabv2} do not generalize well to data from a different domain. In fact, their segmentation performance decreases drastically when there is a discrepancy between the training and test distribution as in a synthetic-to-real scenario. 

The research field of unsupervised domain adaptation (UDA) studies how to transfer knowledge from a labeled source domain to an unlabeled target domain. The aim is to achieve the best possible results in the target domain, whereas the performance in the source domain is not considered. Current methods for UDA address the problem by minimizing the distribution discrepancy between the domains while doing a supervised training on source data. The distribution alignment can take place in the pixel space \cite{CYCADA, Bidirectional, Mobile, FDA}, feature space \cite{FCNs, ActivationMatching, SIM, CAG}, output space \cite{AdaptSegNet, ADVENT, PatchAlign, SIM, CLAN}, or even in several spaces in parallel. 

While adversarial training (AT) \cite{GAN} is commonly used to minimize the distribution discrepancy between domains, it can fail to align the correct semantic categories. This is because adversarial training minimizes the mismatch between global distributions rather than class-specific ones, which can negatively affect the results \cite{SelfGAN}. This is also true for discrepancy measures such as the maximum mean discrepancy \cite{MMD}, which can be minimized without aligning the correct class distributions across domains \cite{CAN}. To address the problem of misaligned classes across domains, we rely on contrastive learning (CL) \cite{ContLoss}. The basic idea of CL is to encourage positive data pairs to be similar and negative data pairs to be apart. To perform domain adaptation and match the features of the correct semantic categories, positive pairs consist of features from the same class but different domains while negative pairs are from different classes and possibly from different domains \cite{CAN, CCSA}. 

Due to the lack of labels in the target domain, the class of each target feature must be determined based on the predictions of the model. However, since different class prior probabilities can bias the final segmentation layer towards the source domain, we extend our approach with self-training (ST), i.e. using target predictions as pseudo-labels. We start from the observation that target pixels are predicted with high uncertainty \cite{ADVENT}. Moreover, these uncertain predictions may also vary between different classes during training and are therefore usually not considered for self-training. By using a memory-efficient temporal ensemble that combines the predictions of a single network over time \cite{TemporalEnsemble}, we obtain the predictive tendency of the model. This allows us to create robust pseudo-labels even for the uncertain predictions that have high information content. The temporal ensemble has the additional advantage that pseudo-labels are updated directly during training, which reduces the computational complexity compared to a separate stage-wise recalculation \cite{CAG, IAST, CCM, CRST, FDA}.

We summarize our contributions as follows: First, we extend contrastive learning and self-training using a memory-efficient temporal ensemble to UDA for semantic segmentation. Second, we empirically show that both approaches are able to transfer knowledge between domains. Furthermore, we show that combining our contrastive learning and self-training (CLST) approach leads to a symbiotic setup that yields competitive and superior state-of-the-art results for GTA5 $\rightarrow$ Cityscapes and SYNTHIA $\rightarrow$ Cityscapes, respectively. 

\section{Related works}

While the focus of UDA was initially on classification, interest in UDA for semantic segmentation has grown rapidly in the last few years. Since this work investigates UDA for semantic segmentation, the following literature review will mainly focus on this topic. In addition, we give a short review on self-ensembling and contrastive learning. 

\textbf{Adversarial Learning:} 
AT can be applied to align the distributions of two domains in pixel space, feature space, and output space. In pixel space, the main idea is to transfer the appearance of one domain to the style of other. Thus, it is assumed that the geometric structure in both domains is approximately the same and the difference is mainly in texture and color. A very common approach uses a CycleGan-based architecture \cite{CycleGAN} to transfer source images to the style of the target domain \cite{CYCADA, Bidirectional, Mobile}. Since the transformation does not change the content, the source labels can be reused to train the model on target-like images in a supervised manner. Similarly, \cite{SelfGAN} uses adaptive instance normalization \cite{AdaIN} in combination with a single Generative Adversarial Network \cite{GAN} to augment source images into different target-like styles. Another option is to use adversarial training to minimize the discrepancy between feature distributions. In this case, the discriminator takes on the role of a domain classifier, which must decide whether the features belong to the source or target domain. For semantic segmentation, this approach was first proposed in \cite{FCNs} and is also used in \cite{CYCADA, Mobile, Proxy}. In \cite{NoMoreDiscrimination}, this approach is extended by additionally giving each class its own domain classifier, which further matches the individual class distributions between domains. \cite{AdaptSegNet, ADVENT, PatchAlign} use adversarial training to align the output spaces on pixel-level and patch level, respectively. Meanwhile, aligning the output distributions with adversarial training is also used in several publications as either a basic component or for warm-up \cite{SIM, MRNet, CAG}.

\textbf{Self-Training}: 
In ST, the target predictions are converted into pseudo-labels, which are then used to minimize the cross-entropy. Since the quality of the pseudo-labels is
crucial to this approach, \cite{FDA} combines the predictions of three models and iteratively repeats the training of the models, followed by the recalculation of the pseudo-labels. Similarly, \cite{MRNet} improves their quality by averaging the predictions of two different outputs of the same network. Another commonly used strategy tries to convert only the correct target predictions into pseudo-labels. \cite{Proxy}, for example, attempts to find them by combining the output of two discriminators with the confidence of a segmentation classifier. \cite{RectifyingPL} does this by considering the pixel-wise uncertainty of the predictions, which is estimated during training. \cite{CAG} converts predictions into pseudo-labels only if the target features are within a certain range of the nearest category-wise source feature mean. \cite{CBST, SIM, CCM, IAST} use the softmax output as confidence measure and incorporate only predictions above a certain threshold into the training process. In doing so, they assume that a higher prediction probability is associated with higher accuracy. In order to make self-training less sensitive to incorrect pseudo-labels, \cite{CRST} uses soft pseudo-labels and smoothed network predictions. Unlike the previously mentioned approaches, \cite{ADVENT} and \cite{MaxSquares} do not explicitly create pseudo-labels but exploit entropy minimization and a maximum squares loss to conduct self-training. Again, both methods can further improve performance when only confident samples are considered.

\textbf{Self-Ensembling:} 
An ensemble considers the outputs of multiple independent models for the same input sample to arrive at a more reliable prediction. The basic idea is that different models make different errors, which can be compensated for in the majority. Ensembling can be employed in ST to create better pseudo-labels for the next training stages \cite{FDA}. In semi-supervised learning, a special variant called self-ensembling has shown remarkable results \cite{TemporalEnsemble, MeanTeacher}. In this case, there is usually only one trainable model that minimizes an additional consistency loss between two different predictions of the same sample. While one prediction remains the output of the trainable network, current methods differ mainly in the creation of the second prediction. \cite{TemporalEnsemble} uses an exponential moving average (EMA) to combine predictions generated at different times during training. This is also known as temporal ensemble. \cite{TemporalEnsemble} also proposes to generate the second prediction with the same model using a different dropout mask and augmentations. \cite{MeanTeacher} extends this idea and proposes a Mean Teacher (MT) framework where there is a second (non-trainable) model whose weights are updated with an EMA over the actual trainable weights. While \cite{SelfWeight} extends the former idea to UDA for classification, \cite{SelfGAN, SelfAttention, DACS} apply the MT framework to UDA for semantic segmentation. 

\textbf{Contrastive Learning:}
CL \cite{ContLoss} is a framework that learns representations by contrasting positive pairs against negative pairs. It has recently dominated the field of self-supervised representation learning \cite{CPC, MoCo, SIMCLR, BigCL}. Contrastive learning encourages representations of positive pairs (typically different augmentations from an image, preserving semantic information) to be similar and negative pairs (different image instances) to be apart.
In the area of domain adaptation (DA) for classification \cite{CAN, CCSA}, CL has been applied to align the feature space of different domains. Positive pairs are generated by using samples of the same class, but different domains. Negative pairs are chosen so that they belong to different classes and potentially different domains. In \cite{CCSA}, the contrastive loss is realized through minimizing a Frobenius norm. \cite{CAN} modifies the maximum mean discrepancy \cite{MMD} to correspond to a contrastive loss. Whereas existing ideas in DA compute the contrastive loss in the feature space, recent work \cite{SIMCLR, BYOL} shows that using a separate projection space for the contrastive loss is very beneficial in the setting of self-supervised representation learning.

\section{Method}
\begin{figure*}[t]
\begin{center}
\def\svgwidth{490pt}
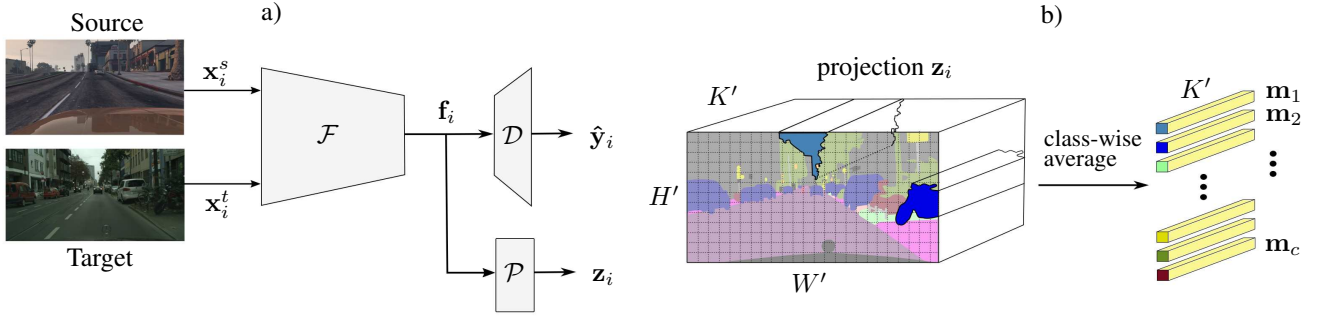
\end{center}
   \caption{a) Network architecture of our proposed approach, consisting of a feature extractor $\mathcal{F}$, a segmentation head $\mathcal{D}$ as well as a projector $\mathcal{P}$. b) Procedure for calculating category-wise mean projections $m_{i,c}$ using a resized segmentation map.}
\label{fig_network}
\end{figure*}

\textbf{Definitions:} In UDA, we have a set $X^s = \{\mathbf{x}_i^s, \mathbf{y}_i^s\}_{i=1}^N$ of $N$ source images $\mathbf{x}_i^s$ with corresponding segmenation maps $\mathbf{y}_i^s$, as well as $M$ unlabeled target images $X^t = \{\mathbf{x}_i^t\}_{i=1}^M$. The indices $s$ and $t$ denote the source and target domain. For simplicity, the image dimensions of both domains are described by $\mathbf{x}_i^s, \mathbf{x}_i^t \in \mathbb{R}^{H\times W \times 3}$. Furthermore, we explicitly divide the network into a feature extractor $\mathcal{F}$ with parameters $\boldsymbol{\theta}_{\mathcal{F}}$ and a segmentation head $\mathcal{D}$ with parameters $\boldsymbol{\theta}_{\mathcal{D}}$. The latter outputs a softmax probability map $\mathbf{p}_i \in \mathbb{R}^{H\times W \times C}$, where $C$ is the total number of classes. The corresponding hard prediction $\mathbf{\hat{y}}_i$ and the source segmentation maps have dimensions $\mathbf{\hat{y}}_i, \mathbf{y}_i^s \in \{0, 1\}^{H\times W \times C}$ and are thus one-hot encoded. As shown in Fig. \ref{fig_network} a), we follow recent findings in self-supervised representation learning and integrate a projector $\mathcal{P}$ with parameters $\boldsymbol{\theta}_{\mathcal{P}}$ into our network \cite{SIMCLR, BYOL}. Its task is to project the extracted feature maps $\mathbf{f}_i \in \mathbb{R}^{H' \times W' \times M'}$ into a projection space $\mathbf{z}_i \in \mathbb{R}^{H' \times W' \times K'}$, where $K'$ is usually smaller than $M'$.  

\subsection{Self-Training} To train the network in a supervised manner, the segmentation maps $\mathbf{y}_i^s$ of the source domain are used to compute a weighted pixel-wise cross-entropy (CE) loss 
\begin{align}
\mathcal{L}_{CE}^s = \frac{-1}{HWC} \sum_{h=1}^H \sum_{w=1}^W \sum_{c=1}^C \alpha_c^s\, \mathbf{y}_i^{s(h,w,c)} \mathrm{log}(\mathbf{p}_i^{s(h,w,c)}),
\label{Eq_ce_loss}
\end{align}
where $\alpha_c^s$ is a class balancing term that will be explained later. Although there is no label information available in the target domain, it is possible to minimize a second CE loss by converting the network predictions into pseudo-labels $\mathbf{\hat{y}}_i^t$. The cross-entropy loss is then given by
\begin{equation}
\mathcal{L}_{CE}^t = \frac{-1}{HWC} \sum_{h=1}^H \sum_{w=1}^W \sum_{c=1}^C \alpha_c^t\, \mathbf{\hat{y}}_i^{t(h,w,c)} \mathrm{log}(\mathbf{p}_i^{t(h,w,c)}), 
\label{Eq_ce_loss_trg}
\end{equation}
where $\alpha_c^t$ is again a weighting term. Minimizing both losses jointly will close the domain gap, if reliable peudo-labels can be provided. Although there may be many noisy predictions, especially in the early stages of the training, it is likely that some pixels will be correctly predicted by the model with some confidence. This is because not all pixels have the same transfer difficulty and some may be easier to transfer than others \cite{CBST, PLCA, CCM, IAST}. To find these pixels during training, we use the entropy of the softmax values
\begin{equation}
\mathbf{e}_i^{t(h, w)} = - \sum_{c=1}^C \mathbf{p}_i^{t(h,w,c)}\, \mathrm{log}(\mathbf{p}_i^{t(h,w,c)}).
\label{Eq_entropy}
\end{equation}
Then, only the predictions that are among the most certain $30\%$ in terms of their entropy are converted into pseudo-labels, while all other predictions are ignored. This approach has already proven successful for MinEnt \cite{ADVENT}. 

Although using the entropy as a guidance for selecting more reliable pseudo-labels works, their quality may  remain moderate because there can be false predictions with high confidence. Therefore, we switch to the following strategy after a few epochs.

\textbf{Temporal Ensemble:} As mentioned earlier, most of the current self-training approaches try to find the correct target predictions by using confidence measures like the entropy or softmax probability. While this strategy can help to avoid including too many false predictions in the training, the information of uncertain but mostly correct predictions is also neglected. Since it is the uncertain forecasts that contain a lot of information, we try to generate reliable pseudo-labels for them as well. This is achieved by considering the predictive tendency of the model for a given sample over time. More precisely, this tendency is extracted by using a temporal ensemble. It considers the numerous target predictions made by the model during training for a specific image. This leads to a smoothing of the noisy predictions. Note that due to the high uncertainty contained in target predictions \cite{ADVENT}, they may even vary between a few classes, especially at the beginning of the training.

For a given image $\mathbf{x}_i^t$, the temporal ensemble is realized through a tensor $\mathbf{t}_i \in \mathbb{N}_0^{H\times W \times C}$ that additively collects the target predictions $\mathbf{\hat{y}}_{i}^t \in \{0, 1\}^{H\times W \times C}$ at time $n$. (Eq. \ref{Eq_ensemble_update}). As the quality of the predictions improves during training, $\mathbf{\hat{y}}_i^t$ is multiplied by a stepwise increasing integer $\gamma_n \in \mathbb{N}$
\begin{equation}
\mathbf{t}_{i,n} \leftarrow \mathbf{t}_{i,n-1} + \gamma_{n}\, \mathbf{\hat{y}}_{i,n}^t. 
\label{Eq_ensemble_update}
\end{equation} This ensures that more recent predictions have a larger impact on the final decision. To generate pseudo-labels from the tensor $\mathbf{t}_i$, majority-voting is used. It selects the class with the most votes as hard prediction $\mathbf{\hat{y}}_i^t$. Although it would be possible to exclude pseudo-labels that have only a slight majority in the temporal ensemble, we do not apply any thresholds here and leave this open to further research. 

The advantage of this realization is that the ensemble can be implemented in a memory-efficient way. First, depending on the number of epochs and $\gamma$, it allows to use the uint8 format for $\mathbf{t}_i$. (In our experiments, $\gamma_n$ did not exceed 4). Second, due to the one-hot encoding of $\mathbf{\hat{y}}_i^t$, the ensemble can be realized as a sparse tensor. This drastically reduces memory requirements when the number of classes $C$ is large. For example, in one of our experiments with $C=19$, only $7.2\%$ of the elements were non-zero. 

\textbf{Class Balancing:}
The motivation for the class balancing terms $\alpha_c^s$ and $\alpha_c^t$ in Eq. \ref{Eq_ce_loss} and Eq. \ref{Eq_ce_loss_trg} arises from the possibility that there may be a discrepancy between the source and target class prior probabilities \cite{PriorAlignment}. Such a difference can bias the final segmentation layer towards the source domain and thus negatively affect the segmentation of target samples \cite{CAG}. To circumvent this problem, we use the source labels and the target pseudo-labels to compute $\alpha_c^s$ and $\alpha_c^t$. Since pseudo-labels can better capture the presence of a class in an image than its exact number of pixels, the following strategy is used for both source and target domain. Instead of computing pixel-wise class prior probabilities, only the occurence probability $o_c^s, o_c^t \in [0, 1]$ of each class $c$ in the dataset is considered. For the source domain, it is given by $o_c^s=N_c\mathbin{/}N$, where $N_c$ is the number of images containing class $c$, while $N$ is the total number of images in the dataset \cite{PriorAlignment}. To save computational resources, $o_c^s$ and $o_c^t$ are approximated online during training starting from probability one for each class. The weighting terms for the source and target domain are then defined by
\begin{align}
\alpha_c(o_c) = \frac{\mathrm{min}(1/o_c, \beta)}{\sum_{i=1}^C \mathrm{min}(1/o_i, \beta)},
\label{Eq_ce_weighting}
\end{align}
where $\beta \geq 1$ is a hyperparamter to avoid a too large weighting of very rare classes. The class balancing terms have the following effects. First, by using $\alpha_c^t$, the model now focuses less on classes that occur frequently in the target dataset. This allows for better knowledge transfer for classes that are rare and perhaps more difficult to transfer \cite{CBST}. Second, by additionally using $\alpha_c^s$, the class prior probabilities of both domains are approximately aligned \cite{PriorAlignment}. 
\begin{figure}[t]
\begin{center}
\def\svgwidth{235pt}
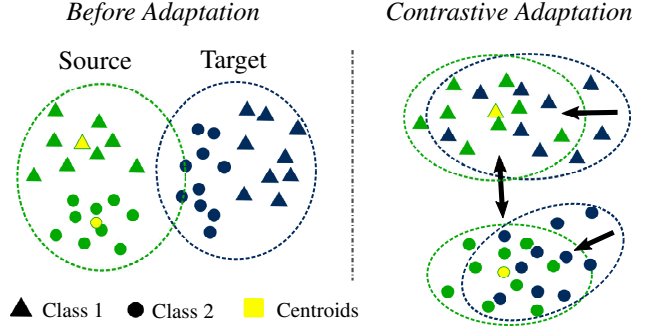
\end{center}
\caption{Left: before adaptation, right: after aligning category-wise centroids across domains using a contrastive loss.}
\label{fig_feature_space}
\end{figure}

\subsection{Contrastive learning} 
Similar to \cite{CAG}, our method is based on the observation that pixels of the same class cluster in the feature space. However, due to the discrepancy between the feature distributions, this observation only applies to features of the source domain and not across domains. This circumstance is also reflected on the left side of Figure \ref{fig_feature_space}. To cluster features of the same class across domains, we use a contrastive loss (right side of Fig. \ref{fig_feature_space}). Following \cite{SIMCLR, BYOL} and in contrast to previous CL approaches proposed for UDA for classification \cite{CAN, CCSA}, the contrastive loss is computed in a projection space $\mathbf{z}_i \in \mathbb{R}^{H' \times W' \times K'}$, see Fig. \ref{fig_network} a). 

First, class-wise mean projections are calculated using all source images in the current batch. For this, it is assumed that the source segmentation masks can be used to assign a class to each projection. Since projections of misclassified features can have an unfavorable effect on the corresponding class means, only the correctly segmented ones are considered. They can be extracted by comparing each prediction with its corresponding ground truth, as in \cite{SIM};
\begin{equation}
\mathbf{\tilde{y}}_i^{s(h,w,c)} = \left\{\begin{matrix}
1 & \mathrm{if}\ \, \mathbf{\hat{y}}_i^{s(h,w,c)} = \mathbf{y}_i^{s(h,w,c)} = 1, \\
0 & \mathrm{otherwise}.
\end{matrix}\right.
\label{Eq_overlapp}
\end{equation} Note that the predictions and labels from Eq. \ref{Eq_overlapp} can have a different hight and width than the source projections $\mathbf{z}_i^s$. In this case, $\mathbf{\hat{y}}_i^s$ and $\mathbf{y}_i^s$ are first resized using nearest neighbor interpolation. Then for each class contained in the current batch, class-wise mean projections are calculated. This is accomplished by averaging over the height $H'$ and width $W'$ of all $B$ source projection maps in the current batch
\begin{align}
\mathbf{m}_c^s = \frac{\sum_{i=1}^B \sum_{h,w} \mathbf{\tilde{y}}_i^{s(h,w,c)}\, \mathbf{z}_i^{s(h,w)}}{\sum_{i=1}^B \sum_{h,w}\mathbf{\tilde{y}}_i^{s(h,w,c)}}, \quad \in \mathbb{R}^{K'}
\label{Eq_centroids_category}
\end{align}
This approach is also illustrated in Fig. \ref{fig_network} b) for a batch consisting of only one source image. While Eq. \ref{Eq_centroids_category} could be calculated using the entire source dataset rather than $B$ samples, this would require propagating all source images through the network once. Furthermore, this procedure would have to be repeated several times, since the projections and thus their mean values will most likely change during training. To still extract a global source centroid $\mathbf{g}_c^s \in \mathbb{R}^{K'}$ for each class $c$ without much computational effort, an exponential moving average is used
\begin{equation}
\mathbf{g}_c^s = (1 - \psi) \mathbf{g}_c^s + \psi \mathbf{m}_c^s,
\label{Eq_global_means}
\end{equation}
where $\psi$ is a momentum term. These global source centroids are also shown in yellow in Fig. \ref{fig_feature_space}. For the target domain, Eq. \ref{Eq_centroids_category} is slightly modified. First, $\mathbf{\tilde{y}}_i^s$ is replaced by (resized) pseudo-labels $\mathbf{\hat{y}}_i^{t} \in \{0, 1\}^{H'\times W'\times C}$. Second, a separate class-wise mean projection $\mathbf{m}_{i,c}^t$ is computed for each target image $\mathbf{x}_i^t$ in the current batch. The reason is that each $\mathbf{m}_{i, c}^t$ should roughly cluster around its respective source centre $\mathbf{g}_c^s$. Finally, the contrastive loss is given by
\begin{equation}
\mathcal{L}_{CL} = - \sum_{c=1}^C \alpha_c^t\, \mathrm{log}\frac{\mathrm{exp}(\mathrm{sim}(\mathbf{g}_c^s, \mathbf{m}_{i, c}^t))}{\sum_{j=1}^C \mathbbm{1}_{[j \neq c]}\ \mathrm{exp}(\mathrm{sim}(\mathbf{g}_j^s, \mathbf{m}_{i,c}^t))}
\label{Eq_contrastive_loss}
\end{equation} 
where $\mathbbm{1}_{[j \neq c]} \in \{0, 1\}$ is the indicator function evaluating to 1 iff $j \neq c$ and $\mathrm{sim(\mathbf{u}, \mathbf{v})} = \mathbf{u}^{\mathrm{T}} \mathbf{v} / \Vert \mathbf{u} \Vert \Vert \mathbf{v} \Vert$ denotes a cosine similarity, which is the dot product between two $l_2$-normalized vectors $\mathbf{u}$ and $\mathbf{v}$. Note that a similar loss was already used in \cite{CPC, SIMCLR, ContLoss1}. It can be shown that $\mathcal{L}_{CL}$ becomes minimal if the cosine similarity becomes maximal for projections of the same class (positive pairs) and minimal for different classes (negative pairs). Thus, the intra-class variance across domains is minimized while the inter-class variance is maximized. This is also denoted by the arrows in Fig. \ref{fig_feature_space}. However, the maximization of the inter-class variance is only implicit because it is the target mean representations $\mathbf{m}_{i, c}^t$ which force the global source centroids $\mathbf{g}_c^s$ to be apart. Since some classes may not always be present in a batch of samples, the contrastive loss is weighted by $\alpha_c^t$ (Eq. \ref{Eq_ce_weighting}). In this case, it encourages the model to focus more on aligning and separating rare classes when they appear in the batch. Finally, note that although a contrastive loss is used, only at most $C^2$ distances need to be calculated for each target image. This makes the approach suitable for models with large feature maps in terms of height and width. 

By combining self-training through temporal ensembling and contrastive learning, we create a symbiotic framework in which both methods can directly benefit each other. ST mitigates the bias of the last segmentation layer towards the source domain improving the target predictions. Needless to say, both ST and CL profit from better pseudo-labels. Additionally, for CL, better pseudo-labels improve the quality of the target class means $\mathbf{m}_{i,c}$. This allows the semantic categories to be clustered more accurately across domains, resulting again in a better target prediction $\mathbf{\hat{y}_i^t}$.

To sum up, the following loss function is minimized
\begin{equation}
\mathcal{L} = \mathcal{L}_{CE}^s + \lambda_{ST} \mathcal{L}_{CE}^t + \lambda_{CL} \mathcal{L}_{CL},
\label{Eq_total_loss}
\end{equation}
where $\lambda_{ST}$ and $\lambda_{CL}$ are two hyperparameters balancing the influence of self-training and contrastive learning. Our overall training procedure is shown in Algorithm \ref{Alg_training}. The derivatives of the losses are computed as $\nabla_{\boldsymbol{\theta}_{\mathcal{F}}} \mathcal{L}, \nabla_{\boldsymbol{\theta}_{\mathcal{D}}} (\mathcal{L}_{CE}^s, \mathcal{L}_{CE}^t)$ and $\nabla_{\boldsymbol{\theta}_{\mathcal{P}}} \mathcal{L}_{CL}$.

\section{Experiments}
\subsection{Experimental Settings}
{\bf Network Architecture:} Similar to \cite{ADVENT, AdaptSegNet, MRNet, SIM, Bidirectional, PLCA, CCM, MaxSquares} we deploy the DeepLab-V2 \cite{Deeplabv2} framework with a ResNet-101 as feature extractor $\mathcal{F}$. As is common practice, we initialize $\mathcal{F}$ with weights pre-trained on ImageNet and freeze all batch normalization layers. The final segmentation head $\mathcal{D}$ consists of an Atrous Spatial Pyramid Pooling (ASPP) head. We fix the dilation rates of the ASPP head to \{6, 12, 18, 24\} as it was done in previous work. To create a symmetrical network architecture, we use a similar ASPP head for the projector $\mathcal{P}$ as described before. The only difference is that the projector outputs a tensor with $K'=256$ channels and does not use any non-linearity. 

\begin{algorithm}
	\caption{CLST Training Procedure}
	\begin{algorithmic}[1]
	\Require{Iterations $T$, Iterations for warm-up $R$,\newline Use ensemble pseudo-labels after $K$ iterations}
	\State Initialize $\boldsymbol{\theta}_{D}$ and $\boldsymbol{\theta}_{P}$ randomly	
	\State Initialize $\boldsymbol{\theta}_{F}$ with ImageNet pre-trained weights
	\State Initialize ensemble tensor $\mathbf{t}$ with zeros
		\For {$j = 1, 2, \dots, T$}
		\State Sample minibatch $\{\mathbf{x}_i^s, \mathbf{y}_i^s\}$, $\{\mathbf{x}_i^t\}$ from $X^s$ and $X^t$
		\State Update source weighting $\alpha_c^s$ using Eq. \ref{Eq_ce_weighting}
		\State Calculate $\mathcal{L}_{CE}^s$  using Eq. \ref{Eq_ce_loss}
		\State Calculate $\mathbf{m}_c^s$ using Eq. \ref{Eq_centroids_category}		
		\State Update $\boldsymbol{\psi}$ using a cosine decay
		\State Update global source centers $\mathbf{g}_c^s$ using Eq. \ref{Eq_global_means}
		\State Get target softmax probability maps $\mathbf{p}_i^t$
		\If {$j > R$}
		\State Update ensemble using Eq. \ref{Eq_ensemble_update} 
		\EndIf
		\If {$j > K$}
		\State Create pseudo-labels from ensemble
		\Else
		\State Generate pseudo-labels for most certain $30\%$
		\EndIf
		\State Update target weighting $\alpha_c^t$ using	Eq. \ref{Eq_ce_weighting}
		\If {$j > R$}
		\State Calculate $\mathcal{L}_{CE}^t$ and $\mathcal{L}_{CL}^t$ using Eq. \ref{Eq_ce_loss_trg} and  \ref{Eq_contrastive_loss}
		\EndIf
		\State Update parameters $\boldsymbol{\theta}_{\mathcal{F}}, \boldsymbol{\theta}_{\mathcal{D}}, \boldsymbol{\theta}_{\mathcal{P}}$
		\EndFor
	\end{algorithmic} 
	\label{Alg_training}
\end{algorithm}

{\bf Implementation Details:}
Our implementation adopts the PyTorch deep learning framework \cite{PyTorch}. Most of the hyperparameters are taken directly from the base architecture from \cite{AdaptSegNet}. We train the network using SGD with Nesterov acceleration to speed up convergence. We deploy a polynomial learning rate decay with an initial learning rate set to $2.5\times10^{-4}$ and an exponent of $0.9$. The momentum of the optimizer is set to $0.9$ and we use a weight decay of $5\times10^{-4}$. Furthermore, we use $0.1$ for both $\lambda_{ST}$ and $\lambda_{CL}$. For the momentum term $\psi$ in Eq. \ref{Eq_global_means} we use a cosine decay starting from $0.02$. For training and testing, we rescale all source images to size $720\times 1280$ and all target images to $512\times1024$ \cite{AdaptSegNet}. In addition, we also investigate the impact of color jittering and Gaussian blur, which was recently used in \cite{DACS}. We train our model using batches with 2 source images and 2 target images. The model is pre-trained for $6$k iterations on the source domain before we switch to the loss function from Eq. \ref{Eq_total_loss}. The pseudo-labels from the temporal ensemble are used after $15$k iterations. $\gamma$ is initially set to 1 and incremented by 1 every $25$k iterations. All iteration-related parameters were chosen to be approximately multiples of the number of target images and received no tuning. We set $\beta$ to $5$, which limits the weighting of classes that occur in less than $20\%$ of the images in the dataset. This was set by inspecting the class occurence probabilities in the GTA5 dataset and was also not tuned.

{\bf Datasets and Metric:} We evaluate our approach in the challenging synthetic-to-real scenario, where Cityscapes \cite{Cityscapes} is used as the real-world target domain. Cityscapes contains 2975 training and 500 validation images with a resolution of $1024 \times 2048$. One of the synthetic source datasets is GTA5 \cite{GTA5}, which contains 24966 synthesized frames of size $1052 \times 1914$ of the well known Grand Theft Auto V video game. We evaluate GTA5 $\rightarrow$ Cityscapes using the common 19 classes. The second synthetic source dataset is SYNTHIA \cite{SYNTHIA}, which has 9400 images in total and only shares 16 classes with Cityscapes. Following \cite{ADVENT, CBST, PLCA, DACS, IAST}, we evaluate this transfer with respect to all 16 classes and for a subset consisting of 13 classes. We train our model using all source images and evaluate it on the Cityscapes validation set. As a metric, we use the widely adopted mean-intersection-over-union (mIoU).

\begin{figure*}[ht!]
\begin{center}
\def\svgwidth{500pt}
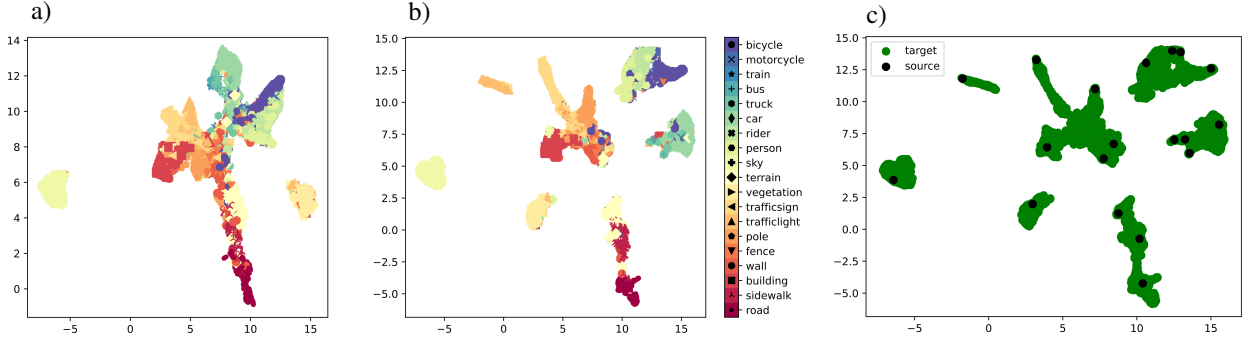
\end{center}
\caption{UMAP feature space visualization of class-wise mean representations for Cityscapes validation set. a) After training on source data only. b) Using our contrastive and self-training approach (CLST). c) CLST with global source centers shown in black.}
\label{fig_UMAP_vis}
\end{figure*}

\subsection{Ablation Studies}
We begin by examining the impact of contrastive learning and self-training on the final results by setting either $\lambda_{ST}$ or $\lambda_{CL}$ to zero. Table \ref{Tab_component_study} a) shows the best mIoU for GTA5 $\rightarrow$ Cityscapes. In addition, it includes the results of the adversarial approach AdaptSegNet \cite{AdaptSegNet} that uses two discriminators to align the output distributions of the segmentation model at different depths. As can be seen, both CL and ST outperform the adversarial baseline and enable knowledge transfer between domains. When CL is combined with ST (CLST), the result improves by $1.7\%$ mIoU compared to plain ST and we reach $49.1\%$  mIoU. This illustrates that CL and ST can indeed benefit from each other.   
\begin{table}[b]
\footnotesize
\setlength\tabcolsep{10pt}
\begin{center}
\begin{tabular}{ l|c|c|c}
  \multicolumn{4}{c}{\small{a) GTA5 $\rightarrow$ Cityscapes}} 
  \\  
  \hline
  Method & $\mathcal{L}_{CL}$ & $\mathcal{L}_{CE}^t$ & mIoU \\
  \hline
 AdaptSegNet \cite{AdaptSegNet} & & & 42.4 \\
 CL & $\checkmark$ & & 43.7 \\
 ST & & $\checkmark$ & 47.4 \\
 CLST & $\checkmark$ & $\checkmark$ & 49.1  \\
 CLST + Aug & $\checkmark$ & $\checkmark$ & $\mathbf{50.6}$  \\
  \hline
\end{tabular}
\end{center}
\end{table}
\begin{table}[b]
\vspace{-4mm}
\footnotesize
\setlength\tabcolsep{7pt}
\begin{center}
\begin{tabular}{ l|c|c|c|c}
\multicolumn{5}{c}{\small{b) SYNTHIA $\rightarrow$ Cityscapes}} 
  \\  
  \hline
  Method & $\mathcal{L}_{CL}$ & $\mathcal{L}_{CE}^t$ & mIoU & mIoU* \\
  \hline
 AdaptSegNet \cite{AdaptSegNet} & & & $-$ & 46.7 \\
 CL & $\checkmark$ & & 40.6 & 46.3 \\
 ST & & $\checkmark$ & 44.9 & 51.7  \\
 CLST & $\checkmark$ & $\checkmark$ & 46.2 & 53.5  \\
 CLST + Aug & $\checkmark$ & $\checkmark$ & $\mathbf{48.5}$ & $\mathbf{56.3}$  \\
  \hline
\end{tabular}
\end{center}
\caption{Component analysis for a) GTA5 $\rightarrow$ Cityscapes and b) SYNTHIA $\rightarrow$ Cityscapes. In the latter case, the results are shown with respect to all 16- and only 13-classes (mIoU*).}
\label{Tab_component_study}
\end{table} To test, why the contribution of contrastive learning is much lower than that of self-training, we examined the influence of the two class balancing terms $\alpha_c^s$ and $\alpha_c^t$. We observed a drop in performance of $1.0\%$ mIoU when only $\alpha_c^t$ was used and more than $3\%$ mIoU when neither $\mathcal{L}_{CE}^s$ nor $\mathcal{L}_{CE}^t$ were weighted. Since contrastive learning cannot mitigate harmful biases in the final segmentation layer towards the source domain, or easy-to-transfer classes, the quality of the pseudo-labels remains moderate. As a result, the target means $\mathbf{m}_{i,c}^t$ include more false predictions, causing a worse alignment. Finally, we added color jittering and Gaussian blur ($+\mathrm{Aug}$), which increased the performance further. 

Similar trends can be observed for SYNTHIA $\rightarrow$ Cityscapes, where the results with respect to all 16- and only 13-classes (mIoU*) are presented in Table \ref{Tab_component_study} b). Again, the contribution of $\mathcal{L}_{CE}^t$ is crucial, but the results can be improved by $1.3\%$ mIoU by adding the contrastive component $\mathcal{L}_{CL}$. In this case, the effect of the augmentations is even larger and an increse of $2.3\%$ mIoU can be observed.

To investigate the influence of our ASPP projector, we calculated the contrastive loss directly on the feature space $\mathbf{f}_i$. As it turned out, the results without projector were around $0.5\%$ mIoU worse for GTA5 $\rightarrow$ Cityscapes. We also experimented with different non-linear projectors having different number of layers and kernels. We found that none of them performed better than our linear ASPP projector, which is symmetrical to our segmentation head and therefore may calculate similar local and global features.

\subsection{Feature Space Visualization}
To visualize the quality of our learned representations, we compute for each target image in the Cityscapes validation set its own category-wise feature mean and visualize all of them using UMAP \cite{UMAP}. This allows us to make a direct comparison with a model trained on source data only. The results are illustrated in Fig. \ref{fig_UMAP_vis}. If a model is trained on source data only a), the clusters are close to each other and are overlapping. As can be seen in b), our CLST algorithm forms much more separable clusters. In c), the global source centroids $\mathbf{g}_c^s$ are now visible in black. Both source and target domain are in similar regions of the feature space, suggesting successful transfer. Note that these experiments were conducted without any data augmentation.

\subsection{Comparisons with state-of-the-art methods}
In Table \ref{Tab_sota_comparison}, we compare our method with the current state-of-the-arts. We mainly show results that also use the DeepLab-V2 with a ResNet-101 as backbone. The only exception to this is CAG \cite{CAG}, which uses the more powerful DeepLabv3+ \cite{Deeplabv3+}. However, since this method minimizes an $l_2$-norm between source and target features and also uses self-training, it was included in the comparison. Furthermore, we cite the results of other approaches directly from the corresponding papers. It is worth mentioning that the cited results may include several training stages \cite{CAG, SIM, MRNet, Bidirectional, CCM, IAST} or transfered source images \cite{SIM, Bidirectional}.
 
\textbf{GTA5 $\rightarrow$ Cityscapes:} 
As shown in Table \ref{Tab_sota_comparison} a), by using similar augmentations like DACS \cite{DACS}, our method (CLST+Aug) beats most of the state-of-the-arts without exiting the training loop to recompute pseudo-labels. While this makes our method significantly less computationally expensive, it comes with slightly worse performance. This is due to the batch normalization layers, which in our experiments give better results in validation mode than in the training mode that is used for the temporal ensemble. If performance is most important, the results can be further improved by creating new pseudo-labels and then fine-tune (FT) the model on the target domain, minimizing only $\mathcal{L}_{CE}^t$ from Eq. \ref{Eq_ce_loss_trg}. This approach was also used by IAST and improves our results by 1$\%$ mIoU for a model that was trained with augmentation (CLST + Aug + FT) and without augmentation (CLST + FT) in the previous stage. Note that the pseudo-labels were created without applying a certainty based threshold. Although other self-training based methods like CAG or IAST may yield competitive results, they rely on an adversarial warm-up. For example, CAG reports a drop of 6.3$\%$ mIoU when  the model was not pre-trained with AT. Our approach, on the other hand, does not require such a warm-up, but may also benefit from it. 

\begin{table*}[t]
\footnotesize
\setlength\tabcolsep{3.5pt}
\begin{center}
\begin{tabular}{ l|ccccccccccccccccccc|c }
  \multicolumn{20}{c}{\small{a) GTA5 $\rightarrow$ Cityscapes}} \\
  \hline
  & \rotatebox[origin=l]{90}{road } & \rotatebox[origin=l]{90}{sidewalk } & \rotatebox[origin=l]{90}{building } & \rotatebox[origin=l]{90}{wall } & \rotatebox[origin=l]{90}{fence } & \rotatebox[origin=l]{90}{pole } & \rotatebox[origin=l]{90}{light } & \rotatebox[origin=l]{90}{sign } & \rotatebox[origin=l]{90}{veg } & \rotatebox[origin=l]{90}{terrain } & \rotatebox[origin=l]{90}{sky } & \rotatebox[origin=l]{90}{person } & \rotatebox[origin=l]{90}{rider } & \rotatebox[origin=l]{90}{car } & \rotatebox[origin=l]{90}{truck } & \rotatebox[origin=l]{90}{bus } & \rotatebox[origin=l]{90}{train } & \rotatebox[origin=l]{90}{motor } & \rotatebox[origin=l]{90}{bike } & mIoU \\
  \hline
 AdaptSeg \cite{AdaptSegNet} & 86.5& 36.0& 79.9& 23.4 & 23.3 & 23.9 & 35.2 & 14.8 & 83.4 & 33.3 & 75.6 & 58.5 & 27.6 & 73.7 & 32.5 & 35.4 & 3.9 & 30.1 & 28.1 & 42.4 \\
 ADVENT \cite{ADVENT} & 89.4 & 33.1 & 81.0 & 26.6 & 26.8 & 27.2 & 33.5 & 24.7 & 83.9 & 36.7 & 78.8 & 58.7 & 30.5 & 84.8 & 38.5 & 44.5 & 1.7 & 31.6 & 32.4 & 45.5 \\
 CBST \cite{CBST} & 91.8 & 53.5 & 80.5 & 32.7 & 21.0 & 34.0 & 28.9 & 20.4 & 83.9 & 34.2 & 80.9 & 53.1 & 24.0 & 82.7 & 30.3 & 35.9 & 16.0 & 25.9 & 42.8 & 45.9 \\
 MaxSquare \cite{MaxSquares} & 89.4 & 43.0 & 82.1 & 30.5 & 21.3 & 30.3 & 34.7 & 24.0 & 85.3 & 39.4 & 78.2 & 63.0 & 22.9 & 84.6 & 36.4 & 43.0 & 5.5 & \textbf{34.7} & 33.5 & 46.4 \\
 PatchAlign \cite{PatchAlign} & 92.3 & 51.9 & 82.1 & 29.2 & 25.1 & 24.5 & 33.8 & 33.0 & 82.4 & 32.8 & 82.2 & 58.6 & 27.2 & 84.3 & 33.4 & 46.3 & 2.2 & 29.5 & 32.3 & 46.5 \\
 PLCA \cite{PLCA} & 84.0 & 30.4 & 82.4 & 35.3 & 24.8 & 32.2 & 36.8 & 24.5 & 85.5 & 37.2 & 78.6 & 66.9 & 32.8 & 85.5 & 40.4 & 48.0 & 8.8 & 29.8 & 41.8 & 47.7 \\
 MRNet \cite{MRNet} & 90.5 & 35.0 & 84.6 & 34.3 & 24.0 & 36.8 & 44.1 & 42.7 & 84.5 & 33.6 & 82.5 & 63.1 & 34.4 & 85.8 & 32.9 & 38.2 & 2.0 & 27.1 & 41.8 & 48.3 \\
 BDL \cite{Bidirectional} & 91.0 & 44.7 & 84.2 & 34.6 & 27.6 & 30.2 & 36.0 & 36.0 & 85.0 & 43.6 & 83.0 & 58.6 & 31.6 & 83.3 & 35.3  & 49.7 & 3.3 & 28.8 & 35.6 & 48.5 \\
 SIM \cite{SIM} & 90.1  & 44.7 & 84.8 & 34.3 & 28.7 & 31.6 & 35.0 & 37.6 & 84.7 & 43.3 & 85.3 & 57.0 & 31.5 & 83.8 & 42.6 & 48.5 & 1.9 & 30.4 & 39.0 & 49.2 \\
 CCM \cite{CCM} & 93.5 & 57.6 & 84.6 & \textbf{39.3} & 24.1 & 25.2 & 35.0 & 17.3 & 85.0 & 40.6 & 86.5 & 58.7 & 28.7 & 85.8 & \textbf{49.0} & \textbf{56.4} & 5.4 & 31.9 & 43.2 & 49.9 \\
 CAG \cite{CAG} & 90.4 & 51.6 & 83.8 & 34.2 & 27.8 & 38.4 & 25.3 & 48.4 & 85.4 & 38.2 & 78.1 & 58.6 & 34.6 & 84.7 & 21.9 & 42.7 & \textbf{41.1} & 29.3 & 37.2 & 50.2 \\
IAST \cite{IAST} & \textbf{93.8} & \textbf{57.8} & 85.1 & 39.5 & 26.7 & 26.2 & 43.1 & 34.7 & 84.9 & 32.9 & 88.0 & 62.6 & 29.0 & \textbf{87.3} & 39.2 & 49.6 & 23.2 & \textbf{34.7} & 39.6 & 51.5 \\
DACS \cite{DACS} & 89.9 & 39.6 & \textbf{87.8} & 30.7 & \textbf{39.5} & \textbf{38.5} & \textbf{46.4} & \textbf{52.7} & \textbf{87.9} & \textbf{43.9} & \textbf{88.7} & \textbf{67.2} & 35.7 & 84.4 & 45.7 & 50.1 & 0.0 & 27.2 & 33.9 & \textbf{52.1}\\
  \hline
 CLST & 90.5 & 42.6 & 83.8 & 35.0 & 26.5 & 24.5 & 40.8 & 35.5 & 84.7 & 37.2 & 81.8 & 63.2 & 36.4 & 85.4 & 41.1 & 51.7 & 0.1 & 25.2 & 47.4 & 49.1 \\
 CLST + FT & 91.2 & 44.5 & 84.4 & 35.9 & 27.4 & 24.0 & 41.2 & 37.3 & 85.3 & 39.7 & 83.1 & 63.7 & 37.6 & 85.9 & 43.2 & 50.8 & 0.1 & 27.7 & 49.7 & 50.1 \\
 CLST + Aug & 92.6 & 52.8 & 85.6 & 35.3 & 27.4 & 28.4 & 42.5 & 37.0 & 84.4 & 36.2 & 87.8 & 63.2 & 35.9 & 86.2 & 43.6 & 49.7 & 0.4 & 25.1 & 48.5 & 50.6 \\
 CLST + Aug + FT & 92.8 & 53.5 & 86.1 & 39.1 & 28.1 & 28.9 & 43.6 & 39.4 & 84.6 & 35.7 & 88.1 & 63.9 & \textbf{38.3} & 86.0 & 41.6 & 50.6 & 0.1 & 30.4 & \textbf{51.7} & 51.6 \\
  \hline
\end{tabular}
\end{center}
\end{table*}

\begin{table*}[t]
\vspace{-3mm}
\footnotesize
\setlength\tabcolsep{4.4pt}
\begin{center}
\begin{tabular}{ l|cccccccccccccccc|cc }
  \multicolumn{19}{c}{\small{b) SYNTHIA $\rightarrow$ Cityscapes}} \\
  \hline
  & \rotatebox[origin=l]{90}{road } & \rotatebox[origin=l]{90}{sidewalk } & \rotatebox[origin=l]{90}{building } & \rotatebox[origin=l]{90}{wall* } & \rotatebox[origin=l]{90}{fence* } & \rotatebox[origin=l]{90}{pole* } & \rotatebox[origin=l]{90}{light } & \rotatebox[origin=l]{90}{sign } & \rotatebox[origin=l]{90}{veg } & \rotatebox[origin=l]{90}{sky } & \rotatebox[origin=l]{90}{person } & \rotatebox[origin=l]{90}{rider } & \rotatebox[origin=l]{90}{car } & \rotatebox[origin=l]{90}{bus } & \rotatebox[origin=l]{90}{motor } & \rotatebox[origin=l]{90}{bike } & mIoU & mIoU* \\
 \hline
 AdaptSeg \cite{AdaptSegNet} & 84.3 & 42.7 & 77.5 & - & - & - & 4.7 & 7.0 & 77.9 & 82.5 & 54.3 & 21.0 & 72.3 & 32.2 & 18.9 & 32.3 & - & 46.7 \\
 ADVENT \cite{ADVENT} & 85.6 & 42.2 & 79.7 & - & - & - & 5.4 & 8.1 & 80.4 & 84.1 & 57.9 & 23.8 & 73.3 & 36.4 & 14.2 & 33.0 & - & 48.0 \\
 CBST \cite{CBST} & 68.0 & 29.9 & 76.3 & 10.8 & 1.4 & 33.9 & 22.8 & 29.5 & 77.6 & 78.3 & 60.6 & 28.3 & 81.6 & 23.5 & 18.8 & 39.8 & 42.6 & 48.9 \\
 MaxSquare \cite{MaxSquares} & 82.9 & 40.7 & 80.3 & 10.2 & 0.8 & 25.8 & 12.8 & 18.2 & 82.5 & 82.2 & 53.1 & 18.0 & 79.0 & 31.4 & 10.4 & 35.6 & 41.4 & 48.2 \\
 PatchAlign \cite{PatchAlign} & 82.4 & 38.0 & 78.6 & 8.7 & 0.6 & 26.0 & 3.9 & 11.1 & 75.5 & 84.6 & 53.5 & 21.6 & 71.4 & 32.6 & 19.3 & 31.7 & 40.0 & 46.5 \\ 
 PLCA \cite{PLCA} & 82.6 & 29.0 & 81.0 & 11.2 & 0.2 & 33.6 & 24.9 & 18.3 & 82.8 & 82.3 & 62.1 & 26.5 & 85.6 & \textbf{48.9} & 26.8 & 52.2 & 46.8 & 54.0 \\
 MRNet \cite{MRNet} & 83.1 & 38.2 & 81.7 & 9.3 & 1.0 & 35.1 & 30.3 & 19.9 & 82.0 & 80.1 & 62.8 & 21.1 & 84.4 & 37.8 & 24.5 & 53.3 & 46.5 & 53.8 \\
 BDL \cite{Bidirectional} & 86.0 & 46.7 & 80.3 & - & - & - & 14.1 & 11.6 & 79.2 & 81.3 & 54.1 & 27.9 & 73.7 & 42.2 & 25.7 & 45.3 & -  & 51.4 \\
 SIM \cite{SIM} & 83.0 & 44.0 & 80.3 & - & - & - & 17.1 & 15.8 & 80.5 & 81.8 & 59.9 & 33.1 & 70.2 & 37.3 & 28.5 & 45.8 & - & 52.1 \\
 CCM \cite{CCM} & 79.6 & 36.4 & 80.6 & 13.3 & 0.3 & 25.5 & 22.4 & 14.9 & 81.8 & 77.4 & 56.8 & 25.9 & 80.7 & 45.3 & 29.9 & 52.0 & 45.2 & 52.9 \\
 CAG \cite{CAG} & 84.7 & 40.8 & 81.7 & 7.8 & 0.0 & 35.1 & 13.3 & 22.7 & \textbf{84.5} & 77.6 & 64.2 & 27.8 & 80.9 & 19.7 & 22.7 & 48.3 & 44.5 & - \\
 IAST \cite{IAST} & 81.9 & 41.5 & \textbf{83.3} & 17.7 & \textbf{4.6} & 32.3 & 30.9 & \textbf{28.8} & 83.4 & 85.0 & 65.5 & 30.8 & \textbf{86.5} & 38.2 & \textbf{33.1} & 52.7 & \textbf{49.8} & 57.0 \\
 DACS \cite{DACS} & 80.5 & 25.1 & 81.9 & \textbf{21.4} & 2.8 & \textbf{37.2} & 22.6 & 23.9 & 83.6 & \textbf{90.7} & \textbf{67.6} & \textbf{38.3} & 82.9 & 38.9 & 28.4 & 47.5 & 48.3 & 54.8 \\
 \hline 
 CLST & 79.0 & 36.7 & 81.3 & 12.7 & 0.3 & 30.4 & 26.3 & 21.5 & 83.7 & 86.2 & 56.4 & 21.1 & 84.6 & 44.8 & 20.7 & 53.4 & 46.2 & 53.5 \\
 CLST + FT & 81.1 & 39.1 & 81.9 & 15.1 & 0.5 & 30.6 & 27.5 & 23.3 & 84.5 & 86.3 & 58.0 & 23.6 & 85.7 & 48.1 & 25.1 & 55.5 & 47.8 & 55.3 \\
 CLST + Aug & 85.9 & 45.7 & 81.9 & 13.8 & 0.3 & 29.6 & 30.5 & 24.2 & 83.6 & 87.9 & 57.7 & 25.3 & 85.2 & 46.9 & 22.3 & 54.8 & 48.5 & 56.3 \\
 CLST + Aug + FT & \textbf{88.0} & \textbf{49.2} & 82.2 & 16.3 & 0.4 & 29.2 & \textbf{31.8} & 23.9 & 84.1 & 88.0 & 59.1 & 27.2 & 85.5 & 46.6 & 28.9 & \textbf{56.5} & \textbf{49.8} & \textbf{57.8} \\
 \hline
\end{tabular}
\end{center}
\caption{Comparison to state-of-the-art results for Cityscapes validation set and task a) GTA5 $\rightarrow$ Cityscapes and b) SYNTHIA $\rightarrow$ Cityscapes. In the latter case, we report the
mIoU with respect to all 16- and only 13-classes, excluding all classes marked with "*".}
\label{Tab_sota_comparison}
\end{table*}

\textbf{SYNTHIA $\rightarrow$ Cityscapes:} Similar results can be observed in Table \ref{Tab_sota_comparison} b), where we show the mIoU with respect to all 16- and only 13-classes, excluding the classes marked with "*". Again, CLST + Aug outperforms most of the other methods, including DACS. If we apply the same strategy as explained before and fine-tune our model on the target domain, we observe an increase of $1.3\%$ mIoU for a model that was trained with augmentation in the first stage. When trained without augmentation, the increase was $1.6\%$ mIoU. Overall, we achieve equivalent results to IAST for 16 classes and better results for 13 classes.

\section{Conclusion}
In this work, we proposed a symbiotic setup for UDA for semantic segmentation. It combines recent ideas in contrastive learning, using a separate projection space, with self-training. For reliable and consistent pseudo-labels over time a memory-efficient temporal ensemble is used. Each individual component already contributes to a knowledge transfer between domains. It is their combination that yields better or equivalent results than the state-of-the-art on two common synthetic-to-real benchmarks: GTA5 $\rightarrow$ Cityscapes and SYNTHIA $\rightarrow$ Cityscapes.

\section*{Acknowledgments}
This publication was created as part of the research project "KI Delta Learning" (project number: 19A19013R) funded by the Federal Ministry for Economic Affairs and Energy (BMWi) on the basis of a decision by the German Bundestag.

{\small
\bibliographystyle{ieee_fullname}
\bibliography{egbib}

\begin{thebibliography}{10}\itemsep=-1pt

\bibitem{Deeplabv2}
Liang-Chieh Chen, George Papandreou, Iasonas Kokkinos, Kevin Murphy, and Alan~L
  Yuille.
\newblock Deeplab: Semantic image segmentation with deep convolutional nets,
  atrous convolution, and fully connected crfs.
\newblock {\em IEEE transactions on pattern analysis and machine intelligence},
  40(4):834--848, 2017.

\bibitem{Deeplabv3+}
Liang-Chieh Chen, Yukun Zhu, George Papandreou, Florian Schroff, and Hartwig
  Adam.
\newblock Encoder-decoder with atrous separable convolution for semantic image
  segmentation.
\newblock In {\em Proceedings of the European conference on computer vision
  (ECCV)}, pages 801--818, 2018.

\bibitem{MaxSquares}
Minghao Chen, Hongyang Xue, and Deng Cai.
\newblock Domain adaptation for semantic segmentation with maximum squares
  loss.
\newblock In {\em Proceedings of the IEEE International Conference on Computer
  Vision}, pages 2090--2099, 2019.

\bibitem{SIMCLR}
Ting Chen, Simon Kornblith, Mohammad Norouzi, and Geoffrey Hinton.
\newblock A simple framework for contrastive learning of visual
  representations.
\newblock {\em arXiv preprint arXiv:2002.05709}, 2020.

\bibitem{BigCL}
Ting Chen, Simon Kornblith, Kevin Swersky, Mohammad Norouzi, and Geoffrey
  Hinton.
\newblock Big self-supervised models are strong semi-supervised learners.
\newblock {\em arXiv preprint arXiv:2006.10029}, 2020.

\bibitem{NoMoreDiscrimination}
Yi-Hsin Chen, Wei-Yu Chen, Yu-Ting Chen, Bo-Cheng Tsai, Yu-Chiang Frank~Wang,
  and Min Sun.
\newblock No more discrimination: Cross city adaptation of road scene
  segmenters.
\newblock In {\em Proceedings of the IEEE International Conference on Computer
  Vision}, pages 1992--2001, 2017.

\bibitem{SelfGAN}
Jaehoon Choi, Taekyung Kim, and Changick Kim.
\newblock Self-ensembling with gan-based data augmentation for domain
  adaptation in semantic segmentation.
\newblock In {\em Proceedings of the IEEE international conference on computer
  vision}, pages 6830--6840, 2019.

\bibitem{Cityscapes}
Marius Cordts, Mohamed Omran, Sebastian Ramos, Timo Rehfeld, Markus Enzweiler,
  Rodrigo Benenson, Uwe Franke, Stefan Roth, and Bernt Schiele.
\newblock The cityscapes dataset for semantic urban scene understanding.
\newblock In {\em Proceedings of the IEEE conference on computer vision and
  pattern recognition}, pages 3213--3223, 2016.

\bibitem{SelfWeight}
Geoffrey French, Michal Mackiewicz, and Mark Fisher.
\newblock Self-ensembling for visual domain adaptation.
\newblock {\em arXiv preprint arXiv:1706.05208}, 2017.

\bibitem{GAN}
Ian Goodfellow, Jean Pouget-Abadie, Mehdi Mirza, Bing Xu, David Warde-Farley,
  Sherjil Ozair, Aaron Courville, and Yoshua Bengio.
\newblock Generative adversarial nets.
\newblock In Z. Ghahramani, M. Welling, C. Cortes, N. Lawrence, and K.~Q.
  Weinberger, editors, {\em Advances in Neural Information Processing Systems},
  volume~27, pages 2672--2680. Curran Associates, Inc., 2014.

\bibitem{BYOL}
Jean-Bastien Grill, Florian Strub, Florent Altch{\'e}, Corentin Tallec,
  Pierre~H Richemond, Elena Buchatskaya, Carl Doersch, Bernardo~Avila Pires,
  Zhaohan~Daniel Guo, Mohammad~Gheshlaghi Azar, et~al.
\newblock Bootstrap your own latent: A new approach to self-supervised
  learning.
\newblock {\em arXiv preprint arXiv:2006.07733}, 2020.

\bibitem{ContLoss}
Raia Hadsell, Sumit Chopra, and Yann LeCun.
\newblock Dimensionality reduction by learning an invariant mapping.
\newblock In {\em 2006 IEEE Computer Society Conference on Computer Vision and
  Pattern Recognition (CVPR'06)}, volume~2, pages 1735--1742. IEEE, 2006.

\bibitem{MoCo}
Kaiming He, Haoqi Fan, Yuxin Wu, Saining Xie, and Ross Girshick.
\newblock Momentum contrast for unsupervised visual representation learning.
\newblock In {\em Proceedings of the IEEE/CVF Conference on Computer Vision and
  Pattern Recognition}, pages 9729--9738, 2020.

\bibitem{CPC}
Olivier Henaff.
\newblock Data-efficient image recognition with contrastive predictive coding.
\newblock In {\em International Conference on Machine Learning}, pages
  4182--4192. PMLR, 2020.

\bibitem{CYCADA}
Judy Hoffman, Eric Tzeng, Taesung Park, Jun-Yan Zhu, Phillip Isola, Kate
  Saenko, Alexei Efros, and Trevor Darrell.
\newblock Cycada: Cycle-consistent adversarial domain adaptation.
\newblock In {\em International conference on machine learning}, pages
  1989--1998. PMLR, 2018.

\bibitem{FCNs}
Judy Hoffman, Dequan Wang, Fisher Yu, and Trevor Darrell.
\newblock Fcns in the wild: Pixel-level adversarial and constraint-based
  adaptation.
\newblock {\em arXiv preprint arXiv:1612.02649}, 2016.

\bibitem{ActivationMatching}
Haoshuo Huang, Qixing Huang, and Philipp Krahenbuhl.
\newblock Domain transfer through deep activation matching.
\newblock In {\em Proceedings of the European Conference on Computer Vision
  (ECCV)}, pages 590--605, 2018.

\bibitem{AdaIN}
Xun Huang and Serge Belongie.
\newblock Arbitrary style transfer in real-time with adaptive instance
  normalization.
\newblock In {\em Proceedings of the IEEE International Conference on Computer
  Vision}, pages 1501--1510, 2017.

\bibitem{CAN}
Guoliang Kang, Lu Jiang, Yi Yang, and Alexander~G Hauptmann.
\newblock Contrastive adaptation network for unsupervised domain adaptation.
\newblock In {\em Proceedings of the IEEE Conference on Computer Vision and
  Pattern Recognition}, pages 4893--4902, 2019.

\bibitem{PLCA}
Guoliang Kang, Yunchao Wei, Yi Yang, Yueting Zhuang, and Alexander~G Hauptmann.
\newblock Pixel-level cycle association: A new perspective for domain adaptive
  semantic segmentation.
\newblock {\em arXiv preprint arXiv:2011.00147}, 2020.

\bibitem{TemporalEnsemble}
Samuli Laine and Timo Aila.
\newblock Temporal ensembling for semi-supervised learning.
\newblock {\em arXiv preprint arXiv:1610.02242}, 2016.

\bibitem{CCM}
Guangrui Li, Guoliang Kang, Wu Liu, Yunchao Wei, and Yi Yang.
\newblock Content-consistent matching for domain adaptive semantic
  segmentation.
\newblock In {\em European Conference on Computer Vision}, pages 440--456.
  Springer, 2020.

\bibitem{Bidirectional}
Yunsheng Li, Lu Yuan, and Nuno Vasconcelos.
\newblock Bidirectional learning for domain adaptation of semantic
  segmentation.
\newblock In {\em Proceedings of the IEEE Conference on Computer Vision and
  Pattern Recognition}, pages 6936--6945, 2019.

\bibitem{CLAN}
Yawei Luo, Liang Zheng, Tao Guan, Junqing Yu, and Yi Yang.
\newblock Taking a closer look at domain shift: Category-level adversaries for
  semantics consistent domain adaptation.
\newblock In {\em Proceedings of the IEEE Conference on Computer Vision and
  Pattern Recognition}, pages 2507--2516, 2019.

\bibitem{UMAP}
Leland McInnes, John Healy, and James Melville.
\newblock Umap: Uniform manifold approximation and projection for dimension
  reduction.
\newblock {\em arXiv preprint arXiv:1802.03426}, 2018.

\bibitem{IAST}
Ke Mei, Chuang Zhu, Jiaqi Zou, and Shanghang Zhang.
\newblock Instance adaptive self-training for unsupervised domain adaptation.
\newblock {\em arXiv preprint arXiv:2008.12197}, 2020.

\bibitem{CCSA}
Saeid Motiian, Marco Piccirilli, Donald~A Adjeroh, and Gianfranco Doretto.
\newblock Unified deep supervised domain adaptation and generalization.
\newblock In {\em Proceedings of the IEEE international conference on computer
  vision}, pages 5715--5725, 2017.

\bibitem{PyTorch}
Adam Paszke, Sam Gross, Soumith Chintala, Gregory Chanan, Edward Yang, Zachary
  DeVito, Zeming Lin, Alban Desmaison, Luca Antiga, and Adam Lerer.
\newblock Automatic differentiation in pytorch.
\newblock 2017.

\bibitem{GTA5}
Stephan~R Richter, Vibhav Vineet, Stefan Roth, and Vladlen Koltun.
\newblock Playing for data: Ground truth from computer games.
\newblock In {\em European conference on computer vision}, pages 102--118.
  Springer, 2016.

\bibitem{SYNTHIA}
German Ros, Laura Sellart, Joanna Materzynska, David Vazquez, and Antonio~M
  Lopez.
\newblock The synthia dataset: A large collection of synthetic images for
  semantic segmentation of urban scenes.
\newblock In {\em Proceedings of the IEEE conference on computer vision and
  pattern recognition}, pages 3234--3243, 2016.

\bibitem{MMD}
Dino Sejdinovic, Bharath Sriperumbudur, Arthur Gretton, and Kenji Fukumizu.
\newblock Equivalence of distance-based and rkhs-based statistics in hypothesis
  testing.
\newblock {\em The Annals of Statistics}, pages 2263--2291, 2013.

\bibitem{Proxy}
Tong Shen, Dong Gong, Wei Zhang, Chunhua Shen, and Tao Mei.
\newblock Regularizing proxies with multi-adversarial training for unsupervised
  domain-adaptive semantic segmentation.
\newblock {\em arXiv preprint arXiv:1907.12282}, 2019.

\bibitem{ContLoss1}
Kihyuk Sohn.
\newblock Improved deep metric learning with multi-class n-pair loss objective.
\newblock In {\em Proceedings of the 30th International Conference on Neural
  Information Processing Systems}, pages 1857--1865, 2016.

\bibitem{HRNet}
Ke Sun, Yang Zhao, Borui Jiang, Tianheng Cheng, Bin Xiao, Dong Liu, Yadong Mu,
  Xinggang Wang, Wenyu Liu, and Jingdong Wang.
\newblock High-resolution representations for labeling pixels and regions.
\newblock {\em arXiv preprint arXiv:1904.04514}, 2019.

\bibitem{MeanTeacher}
Antti Tarvainen and Harri Valpola.
\newblock Mean teachers are better role models: Weight-averaged consistency
  targets improve semi-supervised deep learning results.
\newblock {\em arXiv preprint arXiv:1703.01780}, 2017.

\bibitem{Mobile}
Marco Toldo, Umberto Michieli, Gianluca Agresti, and Pietro Zanuttigh.
\newblock Unsupervised domain adaptation for mobile semantic segmentation based
  on cycle consistency and feature alignment.
\newblock {\em Image and Vision Computing}, 95:103889, 2020.

\bibitem{DACS}
Wilhelm Tranheden, Viktor Olsson, Juliano Pinto, and Lennart Svensson.
\newblock Dacs: Domain adaptation via cross-domain mixed sampling.
\newblock In {\em Proceedings of the IEEE/CVF Winter Conference on Applications
  of Computer Vision}, pages 1379--1389, 2021.

\bibitem{AdaptSegNet}
Yi-Hsuan Tsai, Wei-Chih Hung, Samuel Schulter, Kihyuk Sohn, Ming-Hsuan Yang,
  and Manmohan Chandraker.
\newblock Learning to adapt structured output space for semantic segmentation.
\newblock In {\em Proceedings of the IEEE Conference on Computer Vision and
  Pattern Recognition}, pages 7472--7481, 2018.

\bibitem{PatchAlign}
Yi-Hsuan Tsai, Kihyuk Sohn, Samuel Schulter, and Manmohan Chandraker.
\newblock Domain adaptation for structured output via discriminative patch
  representations.
\newblock In {\em Proceedings of the IEEE International Conference on Computer
  Vision}, pages 1456--1465, 2019.

\bibitem{ADVENT}
Tuan-Hung Vu, Himalaya Jain, Maxime Bucher, Matthieu Cord, and Patrick
  P{\'e}rez.
\newblock Advent: Adversarial entropy minimization for domain adaptation in
  semantic segmentation.
\newblock In {\em Proceedings of the IEEE conference on computer vision and
  pattern recognition}, pages 2517--2526, 2019.

\bibitem{SIM}
Zhonghao Wang, Mo Yu, Yunchao Wei, Rogerio Feris, Jinjun Xiong, Wen-mei Hwu,
  Thomas~S Huang, and Honghui Shi.
\newblock Differential treatment for stuff and things: A simple unsupervised
  domain adaptation method for semantic segmentation.
\newblock In {\em Proceedings of the IEEE/CVF Conference on Computer Vision and
  Pattern Recognition}, pages 12635--12644, 2020.

\bibitem{SelfAttention}
Yonghao Xu, Bo Du, Lefei Zhang, Qian Zhang, Guoli Wang, and Liangpei Zhang.
\newblock Self-ensembling attention networks: Addressing domain shift for
  semantic segmentation.
\newblock In {\em Proceedings of the AAAI Conference on Artificial
  Intelligence}, volume~33, pages 5581--5588, 2019.

\bibitem{PriorAlignment}
Hongliang Yan, Yukang Ding, Peihua Li, Qilong Wang, Yong Xu, and Wangmeng Zuo.
\newblock Mind the class weight bias: Weighted maximum mean discrepancy for
  unsupervised domain adaptation.
\newblock In {\em Proceedings of the IEEE Conference on Computer Vision and
  Pattern Recognition}, pages 2272--2281, 2017.

\bibitem{FDA}
Yanchao Yang and Stefano Soatto.
\newblock Fda: Fourier domain adaptation for semantic segmentation.
\newblock In {\em Proceedings of the IEEE/CVF Conference on Computer Vision and
  Pattern Recognition}, pages 4085--4095, 2020.

\bibitem{CAG}
Qiming Zhang, Jing Zhang, Wei Liu, and Dacheng Tao.
\newblock Category anchor-guided unsupervised domain adaptation for semantic
  segmentation.
\newblock In {\em Advances in Neural Information Processing Systems}, pages
  435--445, 2019.

\bibitem{MRNet}
Zhedong Zheng and Yi Yang.
\newblock Unsupervised scene adaptation with memory regularization in vivo.
\newblock {\em arXiv preprint arXiv:1912.11164}, 2019.

\bibitem{RectifyingPL}
Zhedong Zheng and Yi Yang.
\newblock Rectifying pseudo label learning via uncertainty estimation for
  domain adaptive semantic segmentation.
\newblock {\em International Journal of Computer Vision}, pages 1--15, 2021.

\bibitem{CycleGAN}
Jun-Yan Zhu, Taesung Park, Phillip Isola, and Alexei~A Efros.
\newblock Unpaired image-to-image translation using cycle-consistent
  adversarial networks.
\newblock In {\em Proceedings of the IEEE international conference on computer
  vision}, pages 2223--2232, 2017.

\bibitem{CRST}
Yang Zou, Zhiding Yu, Xiaofeng Liu, BVK Kumar, and Jinsong Wang.
\newblock Confidence regularized self-training.
\newblock In {\em Proceedings of the IEEE International Conference on Computer
  Vision}, pages 5982--5991, 2019.

\bibitem{CBST}
Yang Zou, Zhiding Yu, BVK Vijaya~Kumar, and Jinsong Wang.
\newblock Unsupervised domain adaptation for semantic segmentation via
  class-balanced self-training.
\newblock In {\em Proceedings of the European conference on computer vision
  (ECCV)}, pages 289--305, 2018.

\end{thebibliography}
}

\end{document}